# PoseGRAF: Geometric-Reinforced Adaptive Fusion for Monocular 3D Human Pose Estimation

**Ming Xu, Xu Zhang**

School of Software, Liaoning Technical University, Huludao Campus, Huludao, 125105, Liaoning, China

## Abstract

Existing monocular 3D pose estimation methods primarily rely on joint positional features, while overlooking intrinsic directional and angular correlations within the skeleton. As a result, they often produce implausible poses under joint occlusions or rapid motion changes. To address these challenges, we propose the PoseGRAF framework. We first construct a dual graph convolutional structure that separately processes joint and bone graphs, effectively capturing their local dependencies. A Cross-Attention module is then introduced to model interdependencies between bone directions and joint features. Building upon this, a dynamic fusion module is designed to adaptively integrate both feature types by leveraging the relational dependencies between joints and bones. An improved Transformer encoder is further incorporated in a residual manner to generate the final output. Experimental results on the Human3.6M and MPI-INF-3DHP datasets show that our method exceeds state-of-the-art approaches. Additional evaluations on in-the-wild videos further validate its generalizability. The code is publicly available at https://github.com/iCityLab/PoseGRAF.

## 1 Introduction

Monocular 3D human pose estimation is a fundamental task in computer vision that aims to predict human body poses in 3D space from a single RGB image. It serves as a key enabling technology for a wide range of applications, including motion analysis[1], human-computer interaction[2], [3] and virtual/augmented reality[4] Existing approaches are predominantly divided into two technical paradigms: direct regression of 3D poses from RGB inputs[5], [6] and 2D-to-3D lifting based on detected 2D keypoints[7], [8], [9]. Direct regression methods typically adopt end-to-end convolutional neural networks (CNNs) to estimate 3D poses. In contrast, 2D-to-3D lifting methods first detect 2D keypoints from input images and then infer 3D joint locations. Benefiting from well-established 2D pose detectors, these methods often achieve superior accuracy in practice. Despite the progress, existing 2D-to-3D lifting approaches still face two key limitations: (1) they rely heavily on 2D joint coordinates, which overlooks the underlying structural relationships between joints [10]. (2) they fail to effectively integrate geometric constraints such as bone directions and joint angles [11], [12]. In particular, conventional methods tend to treat bone directions and joint angles as independent limit conditions without modeling their intrinsic correlation, leading to inaccurate pose predictions under complex motion patterns. To address these challenges, we propose PoseGRAF, a novel framework for 3D human pose estimation that integrates geometry-aware graph representation with adaptive feature fusion. PoseGRAF explicitly captures joint angle relationships on the skeleton graph to enhance the representation of bone directions. Specifically, we propose a dual-graph approach to model bone direction relationships, consisting of: (i) a weighted graph, where nodes represent bones and edge weights encode angles between adjacent bones; and (ii) an unweighted graph, where nodes also represent bones and edges indicate binary connectivity (1 for connected, 0 for not connected). Based on this, we design a geometry-enhanced joint embedding method, which integrates Cross-Attention and Joint GCN to extract joint features and employ Bone Direction GCN to integratively encode bone direction and angle information. Furthermore, we design an attention-based dynamic feature fusion module that adaptively fuses positional and geometric features, and co-constructs a residual structure with an improved Transformer encoder. The proposed architecture alleviates unreasonable pose predictions during fast or intricate motions. Extensive experiments on two benchmark datasets, Human3.6M [13] and MPI-INF-3DHP [14] demonstrate that our method outperforms state-of-the-art approaches across multiple metrics, validating its effectiveness and robustness. The main contributions of this work can be summarized as follows:

(1) We design a geometry-enhanced graph to explicitly model the relationships between bone directions and their connections, overcoming the limitations of traditional joint graphs in angle representation. A graph convolution module is designed to effectively capture the spatial correlation of bone directions.

(2) The proposed attention-based dynamic feature fusion module adaptively integrates joint position and bone direction features.

(3) Comprehensive evaluations conducted on the Human3.6M and MPI-INF 3DHP datasets demonstrate that the proposed method achieves superior performance compared to existing state-of-the-art approaches.

## 2 Related work

### 2.1 3D human pose estimation

In recent years, deep learning has significantly advanced the field of monocular 3D human pose estimation. Existing methods can be broadly categorized into two paradigms. The first is direct regression approaches[5], [6], [15] which utilize end-to-end CNN architectures to directly predict 3D joint positions or reconstruct human meshes from raw RGB images. While these methods leverage rich visual information, they are often sensitive to environmental variations and computationally expensive, making them less suitable for real-time processing in dynamic scenes. The second category adopts a two-stage 'image-to-2D-to-3D' pipeline. Chen et al. [16], perform matching and retrieval from a predefined 3D. pose library, achieving computational efficiency but limited by pose diversity. Martinez et al. [17] propose a fully connected residual network that regresses 3D joint positions from 2D keypoints, significantly improving accuracy and benefiting from reliable feature support provided by a detector pretrained on a large-scale 2D dataset. Subsequent improvements, including hierarchical joint prediction [18], keypoint refinement[12], and viewpoint-invariant constraints[19], further enhanced model performance. Although current data augmentation techniques[20] have made significant progress in predictive accuracy, their generalization ability to complex real-world scenarios remains insufficient.

## 2.2 Graph-Based Learning Methods

Graph Convolutional Networks (GCNs) have demonstrated strong performance in monocular 2D-to-3D pose lifting tasks by modeling the topological structure of the human skeleton, where joints are treated as nodes and bones as edges. GCNs can effectively capture spatial dependencies through graph-based convolutional operations. In existing works, Ci et al. [21] proposed a locally connected network to enhance feature representation, SemGCN[22] incorporated joint semantic relationships to refine predictions, and MGCN[23] introduced weight modulation to improve accuracy. However, these approaches rely on static adjacency matrices to define edge weights, making it difficult to model dynamic skeletal interactions. To address this, Zhou et al. [24] proposed Hyperformer, which leverages hypergraph self-attention (HyperSA) to embed skeletal structures into a Transformer framework. While this improves skeletal action recognition, it still falls short in modeling high-order interactions such as bone direction dynamics..

## 2.3 Skeletal Geometry-Aware Methods

Traditional 2D-to-3D pose estimation methods typically regress 3D coordinates directly from 2D joint coordinates. Ma et al.[25] integrated bone length constraints within the GCN framework to mitigate depth ambiguity, while Azizi et al. [26] encoded poses through inter-segment angles to achieve finer skeletal representations. Hu et al. [27] employed a directional graph approach to explicitly model joint-bone relationships, and Yu et al. [28] optimized estimations through GCN-based global-local feature integration. Though these methods emphasize the importance of bone directions and angles, most treat such constraints as auxiliary signals rather than directly incorporating them into graph structures. Sun et al. [10] regressed joint relative displacements through skeletal representations, and Kanazawa et al. [29] incorporated skeletal constraints in 3D mesh reconstruction, yet neither dynamically leveraged angle information. We propose a novel approach: constructing a weighted graph using skeletal orientation angles as edge weights, applied to Graph Convolutional Networks (GCN) processing to achieve higher-accuracy dynamic modeling.

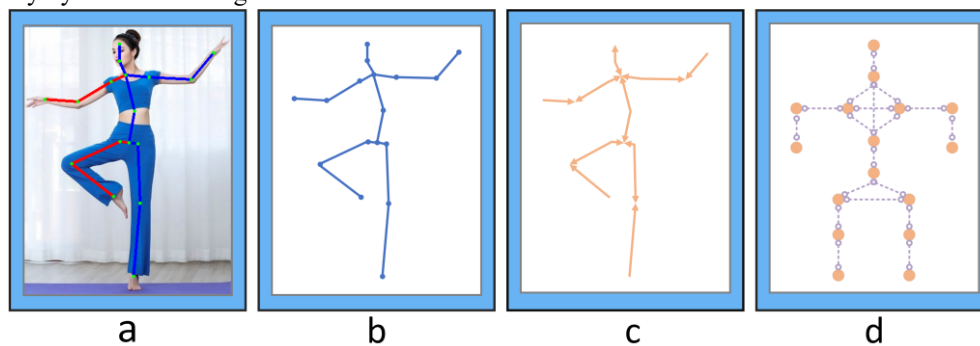

Fig.1. (a) Dance pose. (b) 2D joint graph. (c) Directed weighted bone graph. (d) indicates angles between adjacent bone directions.

# 3. Method

As shown in Fig. 2(a), we propose PoseGRAF, a novel 3D human pose estimation model based on Graph Convolutional Networks (GCN) and Transformer, designed to enhance 3D human pose estimation performance by leveraging advanced graph-based and attention mechanisms to capture the intricate relationships within human skeletal structures.

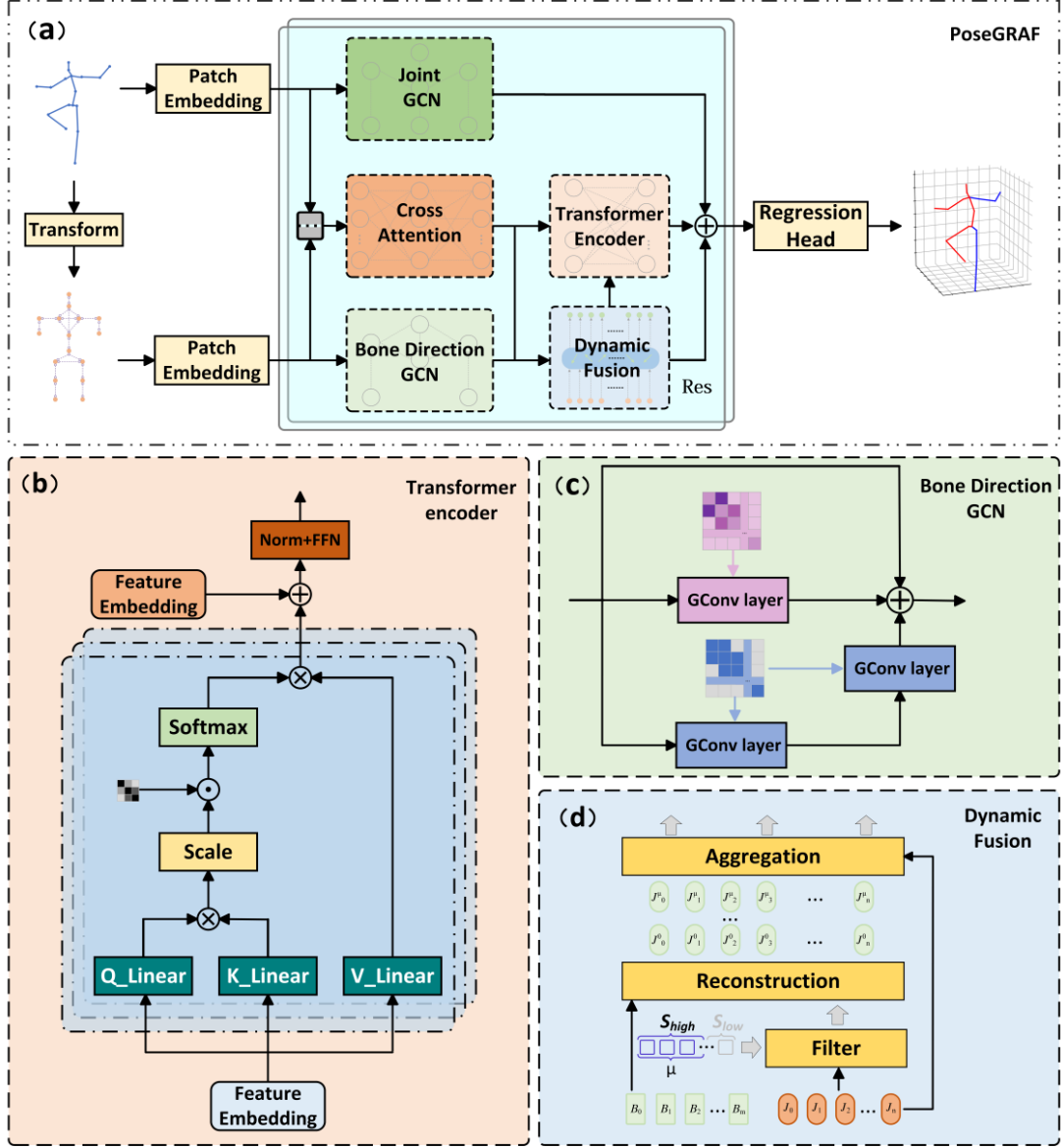

Fig.2. (a) Overview of the proposed framework. ▨ denotes the concatenation of bone directional features and joint features. (b) Transformer encoder module: ▨ represents the relative distance matrix of human body topology, 🟧 indicates feature embeddings processed by Cross-Attention, ⬜ corresponds to embeddings from Dynamic Fusion. (c) Bone-Directional graph convolution module. (d) Dynamic fusion module.

## 3.1 Overview of the network

The PoseGRAF framework consists of five modules: Joint GCN, Bone GCN, Cross-Attention, Dynamic Fusion, and Transformer Encoder. We begin by extracting 2D keypoints from the input image using the CPN detector[30], followed by constructing both directed weighted and undirected graphs to represent skeletal structures. The Bone GCN extracts directional and angular relational features from the skeletal graph. In parallel, the Joint GCN module aggregates features from adjacent nodes to model local spatial dependencies among joints. Next, the Cross-Attention mechanism allows joint features to attend to bone direction features, enhancing the joint representations by incorporating relevant directional information. The Dynamic Fusion module then adaptively integrates the refined joint features with the bone direction features. These fused features are processed by an improved Transformer Encoder, embedded within a residual structure, to generate the final feature representations. Finally, a regression head linearly projects these features to three-dimensional space, enabling accurate estimation of the 3D human pose from the 2D input.

### 3.2 Graph Convolutional Networks

PoseGRAF employs a dual-stream graph convolutional network Architecture comprising Joint GCN and Bone-Direction GCN. The former aims to model local spatial dependencies between human joints, while the latter employs joint angles to generate weighted representations of geometric correlations in bone directions.

**Joint GCN.** We represent joint features as a graph $G_J = (V_J, A_J)$, where the vertex set $V_J$ contains $N$ joints and the edge connections are defined by an adjacency matrix $A_J \in \{0,1\}^{N \times N}$. Specifically, $A_J^{(i,j)} = 1$ if joints $i$ and $j$ share a physical connection, otherwise $A_J^{(i,j)} = 0$. Let $X_J^{(l)}$ denote the latent representation of pose data at the $l-th$ layer, The joint feature representation is updated through graph convolution-based neighbor aggregation, formulated as:

$$X_J^{(l+1)} = \sigma(\widetilde{D}_J^{-1/2} \tilde{A}_J \widetilde{D}_J^{-1/2} X_J^l \Theta_J) \tag{1}$$

Where $\tilde{A}_J = A_J + I_N$ denotes the self–loop augmented adjacency matrix. $\widetilde{D}_J^{-1/2}$ represents the normalized node degree diagonal matrix of $\tilde{A}_J$. $\Theta_J \in \mathbb{R}^{D \times D}$ is a trainable weight matrix.

**Bone Direction GCN.** As shown in Fig. 2(c), this module constructs two geometrically enhanced graphs: a directed weighted bone graph and a directed unweighted bone graph. The directed weighted bone graph is denoted as $G_{BW} = (V_B, W_B)$, where the vertex set $V_B$ contains $M$ bone nodes, constructed as illustrated in Fig. 1. The feature $v_B^p$ of a bone node $x_B^p$ is computed as follows:

$$x_B^p = \frac{x_J^i - x_J^j}{\|x_J^i - x_J^j\|} \tag{2}$$

where $x_J^i$ and $x_J^j$ represent the features of the source joint and target joint of $v_B^p$, respectively. The edge weight between bone nodes $v_B^p$ and $v_B^q$ is computed as follows:

$$w_B^{(p,q)} = \begin{cases} \arccos\left(\dfrac{x_B^p \cdot x_B^q}{\|x_B^p\| \|x_B^p\|}\right), & if\ v_B^p\ and\ v_B^p\ share\ a\ joint \\ 0, & otherwise \end{cases} \tag{3}$$

The directed unweighted bone graph is denoted as $G_{BA} = (V_B, A_B)$, where $A_B^{(p,q)} = 1$, if bone nodes $v_B^p$ and $v_B^q$ share a common joint, otherwise $A_B^{(p,q)} = 0$. Serving as inputs to the Bone Direction GCN module, $G_{BA}$ and $G_{BW}$ undergo feature extraction through two separate graph convolutional layers. These layers update the representations of bone nodes in the subsequent layer by aggregating information from neighboring bone nodes and angular relationships, formulated as follows:

$$\begin{aligned} \bar{X}_W^{(l+1)} &= \sigma(\widetilde{D}_B^{-1/2} \widetilde{W}_B \widetilde{D}_B^{-1/2} X_B^l \Theta_W) \\ \bar{X}_A^{(l+1)} &= \sigma(\widetilde{D}_B^{-1/2} \tilde{A}_B \widetilde{D}_B^{-1/2} X_B^l \Theta_A) \end{aligned} \tag{4}$$

$$X_B^{(l+1)} = \bar{X}_W^{(l+1)} \oplus \bar{X}_A^{(l+1)}$$

Where $\widetilde{W}_B$ denotes the angular-weighted adjacency matrix and $\tilde{A}_B$ represents the original bone connectivity matrix. $\widetilde{D}_B^{-1/2}$ corresponds to the normalized bone node degree diagonal matrix. $\Theta_W, \Theta_A \in \mathbb{R}^{D \times D}$ are two independent learnable parameter matrices. The outputs $\bar{X}_W^{(l+1)}$ and $\bar{X}_A^{(l+1)}$ from the two graph convolutional layers are aggregated through summation to generate the updated node representation features for the $l+1$ layer. The activation function $\sigma(\cdot)$ employs LeakyReLU to mitigate gradient vanishing while preserving feature sparsity, with a negative slope $\alpha=0.01$.

### 3.3 Cross-Attention

This module is designed to capture the intrinsic correlations between human bone directions and joints. We concatenate joint features with bone direction features as follows:

$$X = [X_J^1; X_J^2; \ldots; X_J^N; X_B^1; X_B^2; \ldots; X_B^M] \tag{5}$$

This module takes $X \in \mathbb{R}^{((N+M) \times D)}$ as input, where $D$ denotes the embedding dimension. The module first computes correlation scores between joints and bone directions using the following formulation:

$$\hat{A}_{h;k} = [\hat{a}_{h;k}^{N+1}; \hat{a}_{h;k}^{N+2}; \ldots; \hat{a}_{h;k}^{N+M}] \in \mathbb{R}^{M \times N} \tag{6}$$

Here, $\hat{a}_{h;k}^{N+i}$ denotes the attention score vector between the $i-th$ joint and all bone directions in the $h-th$ head, reflecting the interactions between joints and bone edges. Following [31], we employ Exponential Moving Average (EMA) to aggregate multi-head attention mechanisms across layers.

$$\bar{A}_{h;k} = \beta \cdot \bar{A}_{h;k-1} + (1-\beta) \cdot \bar{A}_{h;k} \tag{7}$$

where $\beta=0.99$. The final layer's $\bar{A}_{h;k}$ is then employed to aggregate attention vectors from different heads across joints and bone directions, yielding the final visual token correlation scores:

$$S = \frac{1}{HM} \sum_{h=1}^{H} \sum_{i=1}^{M} \bar{a}_{h;k}^{(N+i)} \tag{8}$$

where $\bar{a}_{h;k}^{N+i}$ represents the $i-th$ column of matrix $\bar{A}_{h;k}$, with $H$ denoting the number of attention heads. After cross-attention processing, the output features are partitioned into bone direction features $X_{BC} \in \mathbb{R}^{M \times D}$ and joint features $X_{JC} \in \mathbb{R}^{N \times D}$, enabling independent processing by subsequent modules. This design not only preserves the structural information of features but also explicitly models multi-scale dependencies between key joints and bone directions through the multi-head attention mechanism. By incorporating attention mechanisms, this module significantly enhances the model's capability to capture relationships between critical joints and bone directions in human poses, thereby providing richer and more precise feature representations for downstream pose estimation tasks.

### 3.4 Dynamic_Fusion

Inspired by An et al.[32], we propose an attention-based dynamic feature fusion mechanism. This mechanism effectively fuses joint feature embeddings with bone direction embeddings, as illustrated in Fig. 2(d). The Feature Selection function Filter implements adaptive key joint selection through a learnable threshold parameter $\mu$, balancing computational efficiency with precision. Where $S_{high}$ denotes attention scores, based on which the $top-\mu$ joint features with the strongest skeletal correlations are extracted. This process is a dynamic feature gating operation to learn feature subset selection. The graph reconstruction function Reconstruction restores global joint features from individual joint node features according to All the Bone Direction Features, with its detailed implementation described in Algorithm 1 (lines 5-7). This function achieves feature reconstruction through iterative topological diffusion. Specifically, starting from a single selected key joint feature $X_{JC}^{(i)}$ as propagation seeds, a Breadth-First Search (BFS) is performed based on the topological structure of human skeletal graph $G_J$. While in incorporating bone direction features $X_B$ during traversal. The process is mathematically formulated as:

$$\mathcal{J}_B^i = BFS(G_J, X_{JC}^{(i)}, X_B) \tag{9}$$

$\mathcal{J}_B^i$ represents the global joint feature obtained from the joint feature $X_{JC}^{(i)}$. The BFS process takes as input and encodes joint features based on bone direction features $X_B$. The final joint descriptor features are obtained through aggregation and residual connections. This process is formulated as in Eq. (10):

$$X_{DF} = \sum_{i=0}^{\mu-1} \mathcal{J}_B^i + X_{JC} \tag{10}$$

Where the $X_{JC}$ preserves original joint information to prevent gradient vanishing. By integrating bone direction features with joint features obtained through attention mechanisms, this module enhances the spatial representational capacity of each joint. Through this approach, the model can accurately capture geometric relationships between joints and prioritize key points via attention mechanisms.

**Algorithm 1:** Dynamic Fusion
**Input:** $X_{JC}, X_B, S_{high}, G_J$
**Output:** Fused feature $X_{DF}$
1. \\Filter
2. $Index = Top\_Indices(S_{high}, X_{JC})$
3. $\mathcal{X}_{JC} = \{X_{JC}^i | i \in Index\}$
4. \\ Reconstruction
5. for $X_{JC}^i \in \mathcal{X}_{JC}$ do
6. $\quad \mathcal{J}_B^i = BFS(G_J, X_{JC}^i, X_B)$
7. $\quad \mathcal{J}_B.add(\mathcal{J}_B^i)$
8. end for
9. $X_{DF} = \sum_{i=0}^{\mu-1} \mathcal{J}_B^i + X_{JC}$

### 3.5 Transformer Encoder

The conventional Transformer encoder can model global dependencies through multi-head self-attention, enabling each node to equally influence others. However, its permutation-invariant property neglects the critical topological inductive bias in human pose estimation, where interactions between anatomically adjacent joints are inherently stronger than those between distant nodes, thus requiring explicit encoding of structural relationships [33]. Existing graph positional encoding methods fail to effectively perceive node distances. Inspired by [34], as shown in Fig. 2(b), we introduce an enhanced Transformer encoder specifically for 2D-to-3D lifting in human pose estimation, incorporating a relative distance matrix derived from human topology to regulate attention preferences toward distant nodes. We rescale the human joint topology matrix $A_J$ to adjust attention weights:

$$\dot{A} = \frac{1+\exp(w)}{1+\exp(w-A_J)} \tag{11}$$

The hyperparameter $w$ controls the distance−aware information intensity, where larger $w$ values prioritize denser information from distant nodes. In this work, we set $w$ to facilitate information exchange between non-local nodes while maintaining balanced interactions.

To preserve original global joint features, the joint structural information extracted by the Cross_Attention module is injected into the multi-head attention layers of the transformer encoder through residual connections, formulated as:

$$X_{mid} = softmax(\frac{ReLU(Q_g^{(i)}K_g^{(i)T})\odot \dot{A}}{\sqrt{d_g}})v_g^{(i)} + X_{JC} \tag{12}$$

The symbol $\odot$ denotes element-wise product, $1/\sqrt{d_g}$ is the attention scaling factor where $d_g$ represents the dimensionality of vectors in $K_g^{(i)}$. The FFN is then applied to $X_{mid}$ to generate the output of the Transformer encoder.

$$FFN(X) = \sigma(\sigma(\sigma(X_{mid}W_1 + b_1)W_2 + b_2)W_3 + b_3 \tag{13}$$

Here $W_1$, $W_2$ and $W_3$ are trainable matrices, while $b_1$, $b_2$ and $b_3$ denote bias terms. For the activation function $\sigma$, we employ the Gaussian Error Linear Unit (GELU).

### 3.6 Loss function

During model training, we adopt the Mean Per Joint Position Error (MPJPE) as the optimization objective function. This metric optimizes the model by computing the mean Euclidean distance between predicted and ground-truth joint positions in 3D pose space for all joints. Its mathematical expression is defined as:

$$\mathcal{L} = \frac{1}{ZN}\sum_{i=1}^{Z}\sum_{j=1}^{N}\|Y^{(i,j)} - \hat{Y}^{(i,j)}\|_2 \tag{14}$$

Here, $Y^{(i,j)}$ denotes the annotated 3D coordinates of the $j$-th joint in the $i$-th sample, $\hat{Y}^{(i,j)}$ represents the predicted coordinates of the corresponding joint output by the network, $Z$ is the batch size, and $N$ indicates the total number of human joints.

Table 1. Experimental comparisons on the Human3.6M dataset use 2D poses detected by CPN as network input. The symbol (&) indicates models utilizing temporal information, the symbol (*) denotes models employing sharpness-optimized input processing, and best results are highlighted in bold.

| MPJPE(mm)↓ | Dir. | Disc. | Eat. | Greet. | Phone. | Photo. | Pose. | Purch. | Sit. | SitD. | Smoke. | Wait. | WalkD. | Walk | WalkT. | Avg. |
|---|---|---|---|---|---|---|---|---|---|---|---|---|---|---|---|---|
| GraphSH[35] | 45.2 | 49.9 | 47.5 | 50.9 | 54.9 | 66.1 | 48.5 | 46.3 | 59.7 | 71.5 | 51.4 | 48.6 | 53.9 | 39.9 | 44.1 | 51.9 |
| PoseFormer&[9] | 46.9 | 51.9 | 46.9 | 51.2 | 53.4 | 60.0 | 49.0 | 47.5 | 58.8 | 67.2 | 51.6 | 48.9 | 54.3 | 40.2 | 42.1 | 51.3 |
| GraFormer[36] | 45.2 | 50.8 | 48.0 | 50.0 | 54.9 | 65.0 | 48.2 | 47.1 | 60.2 | 70.0 | 51.6 | 48.7 | 54.1 | 39.7 | 43.1 | 51.8 |
| MixSTE&[37] | 46.0 | 49.9 | 49.1 | 50.8 | 52.7 | 58.4 | 48.4 | 47.3 | 60.3 | 67.6 | 51.4 | 48.5 | 53.8 | 39.5 | 42.7 | 51.1 |
| UGRN[38] | 47.9 | 50.0 | 47.1 | 51.3 | 51.2 | 59.5 | 48.7 | 46.9 | 56.0 | 61.9 | 51.1 | 48.9 | 54.3 | 40.0 | 42.9 | 50.5 |
| RS-Net*[39] | 44.7 | **48.4** | 44.8 | 49.7 | **49.6** | 58.2 | 47.4 | 44.8 | 55.2 | 59.7 | 49.3 | 46.4 | 51.4 | 38.6 | 40.6 | 48.6 |
| DGFormer[8] | 46.3 | 50.3 | 45.7 | 50.5 | 50.8 | 57.5 | 49.6 | 46.0 | 55.8 | 63.8 | 50.9 | 47.8 | 53.0 | 38.7 | 41.3 | 49.8 |
| GraphMLP[40] | 45.4 | 50.2 | 45.8 | 49.2 | 51.6 | 57.9 | 47.3 | 44.9 | 56.9 | 61.0 | 49.5 | 46.9 | 53.2 | 37.8 | 39.9 | 49.2 |
| Ours | **44.1** | 49.1 | **43.5** | **48.0** | 50.1 | **56.3** | **46.3** | **44.1** | 56.9 | 60.1 | **49.3** | **45.7** | **51.3** | **36.8** | **39.6** | **48.1** |
| P-MPJPE(mm)↓ | Dir. | Disc. | Eat. | Greet. | Phone. | Photo. | Pose. | Purch. | Sit. | SitD. | Smoke. | Wait. | WalkD. | Walk | WalkT. | Avg. |
| SRNet[7] | 35.8 | 39.2 | 36.6 | 36.9 | 39.8 | 45.1 | 38.4 | 36.9 | 47.7 | 54.4 | 38.6 | 36.3 | 39.4 | 30.3 | 35.4 | 39.4 |
| Liu et al. [41] | 35.9 | 40.0 | 38.0 | 41.5 | 42.5 | 51.4 | 37.8 | 36.0 | 48.6 | 56.6 | 41.8 | 38.3 | 42.7 | 31.7 | 36.2 | 41.2 |
| MGCN*[23] | 35.7 | 38.6 | 36.3 | 40.5 | 39.2 | 44.5 | 37.0 | 35.4 | 46.4 | 51.2 | 40.5 | 35.6 | 41.7 | 30.7 | 33.9 | 39.1 |
| SGNN[42] | 33.9 | 37.2 | 36.8 | 38.1 | 38.7 | 43.5 | 37.8 | 35.0 | 47.2 | 53.8 | 40.7 | 38.3 | 41.8 | 30.1 | 31.4 | 39.0 |
| RS-Net*[39] | 35.5 | 38.3 | 36.1 | 40.5 | 39.2 | 44.8 | 37.1 | 34.9 | **45.0** | 49.1 | 40.2 | 35.4 | 41.5 | 31.0 | 34.3 | 38.9 |
| HopFIR[43] | **43.9** | **47.6** | **45.5** | 48.9 | 50.1 | 58.0 | **46.2** | 44.5 | 55.7 | 62.9 | **49.0** | 45.8 | 51.8 | 38.0 | 39.9 | 48.5 |
| DGFormer[8] | 35.4 | 38.4 | 35.8 | 40.3 | 39.2 | 43.7 | 37.6 | 34.8 | **44.7** | 51.3 | 40.2 | 36.1 | 41.2 | 30.6 | 33.9 | 38.9 |
| GraphMLP[40] | 35.0 | 38.4 | 36.6 | **39.7** | 40.1 | 43.9 | **35.9** | 34.1 | 45.9 | 48.6 | **40.0** | 35.3 | 41.6 | 30.0 | 33.3 | 38.6 |
| Ours | **34.9** | **38.0** | **35.5** | 39.8 | **40.1** | **43.3** | 36.3 | **33.8** | 46.4 | **48.6** | 40.5 | **35.3** | **40.8** | **29.7** | **33.4** | **38.3** |

## 4. Experiments

### 4.1 Datasets and evaluation metrics

This section presents comprehensive studies on two real-world 3D human pose estimation benchmark datasets to systematically validate the superiority of the proposed model.

**Human3.6M Datasets:** As the most representative benchmark in 3D human pose estimation, the Human3.6M Dataset [13] provides 3.6 million frames of multi-view motion captured data captured by four synchronized cameras at a 50 Hz sampling rate, covering 15 categories of daily activities performed by 11 subjects in indoor scenes. Following the standard experimental protocol, we adopt data from subjects (S1, S5, S6, S7, S8) for model training, and evaluate performance on two subjects (S9, S11). Two mainstream evaluation metrics are employed: Protocol 1 (MPJPE) measures absolute errors by computing the Euclidean distance (in millimeters) between predicted and ground-truth 3D joint coordinates; Protocol 2 (P-MPJPE) calculates relative errors after aligning predictions with ground truth via Procrustes analysis.

**MPI-INF-3DHP Dataset:** The MPI-INF-3DHP Dataset [14] is a more challenging 3D human pose estimation benchmark, capturing 1.3 million frames of diverse poses from 8 subjects in indoor/outdoor hybrid scenes using 14 cameras. Aligned with settings in [11], [9], and [14], we utilize the Percentage of Correct Keypoints (PCK) under a 150 mm radius and the Area Under the Curve (AUC) as evaluation metrics.

Table 2. presents experimental comparisons on the Human3.6M dataset using ground-truth 2D poses as network input. The symbol (*) indicates models utilizing temporal information. Best results are highlighted in bold.

| MPJPE(mm)(↓) | Dir. | Disc. | Eat. | Greet. | Phone. | Photo. | Pose. | Purch. | Sit. | SitD. | Smoke. | Wait. | WalkD. | Walk | WalkT. | Avg. |
|---|---|---|---|---|---|---|---|---|---|---|---|---|---|---|---|---|
| Liu et al [41] | 36.8 | 40.3 | 33.0 | 36.3 | 37.5 | 45.0 | 39.7 | 34.9 | 40.3 | 47.7 | 37.4 | 38.5 | 38.6 | 29.6 | 32.0 | 37.8 |
| SRNet [7] | 35.9 | 36.7 | 29.3 | 34.5 | 36.0 | 42.8 | 37.7 | 31.7 | 40.1 | 44.3 | 35.8 | 37.2 | 36.2 | 33.7 | 34.0 | 36.4 |
| PoseGTAC [44] | 37.2 | 42.2 | 32.6 | 38.6 | 38.0 | 44.0 | 40.7 | 35.2 | 41.0 | 45.5 | 38.2 | 39.5 | 38.2 | 29.8 | 33.0 | 38.2 |
| GraphSH [35] | 35.8 | 38.1 | 31.0 | 35.3 | 35.8 | 43.2 | 37.3 | 31.7 | 38.4 | 45.5 | 35.4 | 36.7 | 36.8 | 27.9 | 30.7 | 35.8 |
| GraFormer [36] | 32.0 | 38.0 | 30.4 | 34.4 | 34.7 | 43.3 | 35.2 | 31.4 | 38.0 | 46.2 | 34.2 | 35.7 | 36.1 | 27.4 | 30.6 | 35.2 |
| PHGANet [45] | 32.4 | 36.5 | 30.1 | 33.3 | 36.3 | 43.5 | 36.1 | 30.5 | 37.5 | 45.3 | 33.8 | 35.1 | 35.3 | 27.5 | 30.2 | 34.9 |
| DGformer [8] | 31.5 | 34.3 | 28.2 | 32.2 | 31.3 | 36.8 | 37.0 | 29.4 | 34.9 | 37.8 | 31.8 | 32.5 | 33.0 | 26.7 | 28.9 | 32.4 |
| GraphMLP [40] | 32.2 | 38.2 | 29.3 | 33.4 | 33.5 | 38.1 | 38.2 | 31.7 | 37.3 | 38.5 | 34.2 | 36.1 | 35.5 | 28.0 | 29.3 | 34.2 |
| Ours | 30.9 | 35.5 | 27.2 | 31.6 | 31.7 | 36.3 | 36.4 | 30.3 | 36.6 | 35.0 | 31.4 | 34.3 | 32.5 | 25.6 | 26.2 | 32.1 |

## 4.2 Implementation details

Our method is implemented using PyTorch on a single NVIDIA RTX 3090 GPU. Core architectural parameters are configured as follows: the Transformer encoder comprises L = 6 stacked layers, each self-attention layer contains h = 8 attention heads, and the feature embedding dimension is d = 512. During training, horizontal flipping data augmentation is applied to enhance model robustness, while the same strategy is synchronized in the test phase for result ensembling. The optimization process employs the Adam optimizer with an initial learning rate of 0.001 and an exponential decay scheduler (decay rate $\gamma$ = 0.96), and is trained for 40 epochs. For 2D pose detection, both Human3.6M and MPI-INF-3DHP datasets utilize the Cascaded Pyramid Network (CPN) [30] as the base detector to ensure reliable 2D input features.

## 4.3 Comparsion with state-of-the art

**Result On Human3.6M:** As shown in Tables 1 and 2, when using 2D poses detected by CPN as input, our model outperforms existing methods on both MPJPE (48.1 mm) and P-MPJPE (38.3 mm) metrics. Compared to state-of-the-art graph transformer approaches, PoseGRAF achieves a reduction of 10.6 mm in MPJPE over GraFormer[36] and 1.1 mm over GraphMLP[40]. Notably, PoseGRAF demonstrates superior 3D pose prediction accuracy in complex motion scenarios such as Phoning and Walking. Quantitative analysis reveals that the fusion of geometric features—joint positions, bone directions, and joint angles—significantly improves pose estimation accuracy, effectively enhancing geometric consistency between predictions and ground-truth annotations.

**Result on MPI-INF-3DHP:** We further validate the generalization capability of our model PoseGRAF using the MPI-INF-3DHP dataset, which contains diverse pose variations. The model trained on Human3.6M is directly applied to regress 3D pose coordinates. As shown in Table 3, our method achieves state-of-the-art performance on both PCK and AUC metrics. These results demonstrate that the proposed model exhibits strong generalization and effectively adapts to unseen data.

Table. 3. Results on MPI-INF-3DHP

| Method | Outdoor | PCK(%)(↑) | AUC(%)(↑) |
| --- | --- | --- | --- |
| SGNN[42] | 84.6 | 82.1 | 46.2 |
| MGCN[23] | 85.7 | 86.1 | 53.7 |
| GraphSH[35] | - | 76.4 | 39.3 |
| PoseGTAC[44] | - | 80.1 | 45.8 |
| GraFormer[36] | 74.1 | 79.0 | 43.8 |
| UGRN[38] | 81.9 | 84.1 | 53.7 |
| RS-Net[39] | - | 85.6 | 53.2 |
| DGFormer[8] | - | 85.5 | 53.6 |
| GraphMLP[40] | 86.3 | 87.0 | 54.3 |
| Ours | 87.1 | 88.3 | 54.8 |

Table 4. Ablation studies on Human3.6M with ground truth 2D poses as network inputs.

| Method | MPJPE | P-MPJPE |
| --- | --- | --- |
| Transformer+J-GCN (Baseline) | 34.6 | 27.1 |
| Transformer+ J-GCN +B-GCN+S-Fusion | 33.7 | 26.6 |
| Transformer+J-GCN+Attention | 34.0 | 26.8 |
| Transformer+J-GCN+B-GCN+S-Fusion + C-Attention | 32.8 | 25.3 |
| Transformer+J-GCN+B-GCN+D-Fusion+ C-Attention | 32.1 | 25.0 |

**Qualitative Results:** Fig. 4. compares the prediction results of PoseGRAF, GraphMLP[40], and baseline models against ground-truth poses on representative samples from both datasets. Observations of key regions marked with green and purple circles reveal that PoseGRAF consistently outperforms baseline models and GraphMLP in pose prediction accuracy, regardless of pose complexity. Notably, PoseGRAF maintains precise 3D pose estimation even in highly dynamic motion scenarios, attributed to the dynamic fusion mechanism's effective modeling of geometric relationships in poses.

## 4.4 Ablation studies

To comprehensively evaluate the effectiveness of model components, this study conducts systematic ablation experiments on the Human3.6M dataset. The experimental design includes validation of module effectiveness and model depth optimization. All experiments employ ground-truth 2D pose inputs to exclude interference from detection errors.

**Baseline Model:** The baseline model consists of a Transformer encoder (6-layer × 8-head configuration) cascaded with a Joint GCN module, with a fixed feature embedding dimension of 512.

**Model Depth Optimization:** As shown in Fig. 3, the Transformer encoder depth(L) and feature dimension(D) exhibit significant impacts on model performance: When $L = 6$, MPJPE decreases linearly with increasing layers (reaching the optimal value of 48.1 mm at $L = 6$), but performance degrades when $L > 6$ due to gradient propagation attenuation (error rebounds to 51.3 mm at $L = 8$). Analysis of feature dimensions indicates that $D = 512$ substantially enhances model capacity compared to $D = 256$.

Consequently, the configuration L = 6 and D = 512 is selected as the optimal setup.

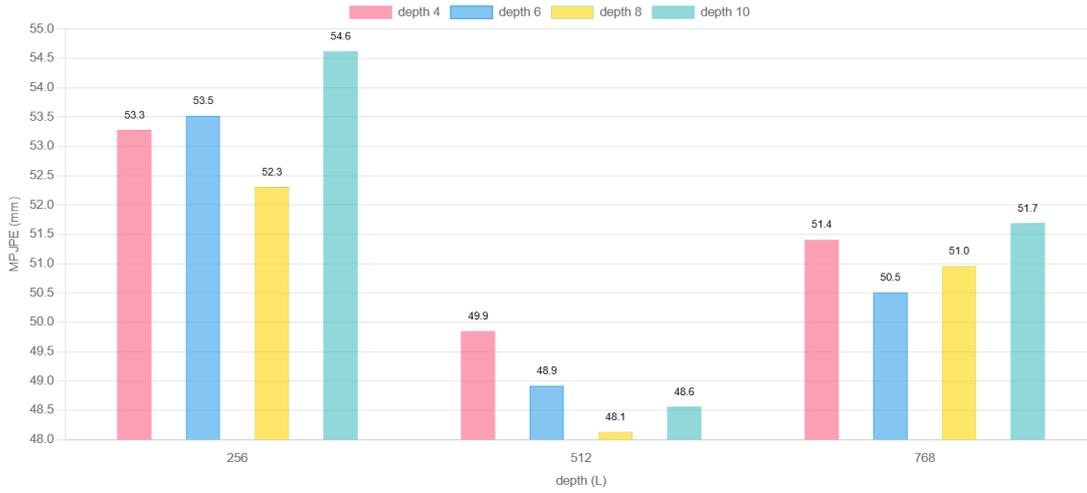

Fig. 3. Architecture Parameter Analysis (Depth, Dimensions) in PoseGRAF. Evaluated on Human3.6M using MPJPE (mm) with CPN-detected 2D poses as network inputs.

**Module Effectiveness Validation**: Through systematic ablation studies (Table 4), we rigorously quantify the individual contributions of the dynamic feature fusion mechanism and the bone direction graph convolutional network (B-GCN). We analyze their effects on 2D-to-3D pose estimation through comparative experiments. The dynamic fusion module constructs an association weight matrix between joint features and bone direction features via Cross-Attention, dynamically selecting critical feature subsets (Top-µ). Compared to static fusion (which directly fuses all 17 joint features and increases MPJPE by 4.1 mm), dynamic fusion improves salient feature selection through attention weights, suppresses redundant interference, and significantly reduces joint localization errors under complex motions. Experiments further verify the role of B-GCN. The baseline model (Transformer + J-GCN) achieves an MPJPE of 34.6 mm. Integrating B-GCN with static fusion reduces errors to 33.7 mm (MPJPE) and 26.60 mm (P-MPJPE), demonstrating that explicit bone direction modeling enhances geometric constraints. With additional Cross-Attention integration, MPJPE further decreases to 32.86 mm. Finally, the complete model with dynamic fusion (MPJPE=32.1 mm, P-MPJPE=25.0 mm) achieves a 7.2% error reduction over the baseline, attributed to its dual-path design: dynamic weighted graphs adaptively adjust feature association strength to precisely capture local motion patterns, while static adjacency graphs encode anatomical priors to reinforce skeletal connectivity.

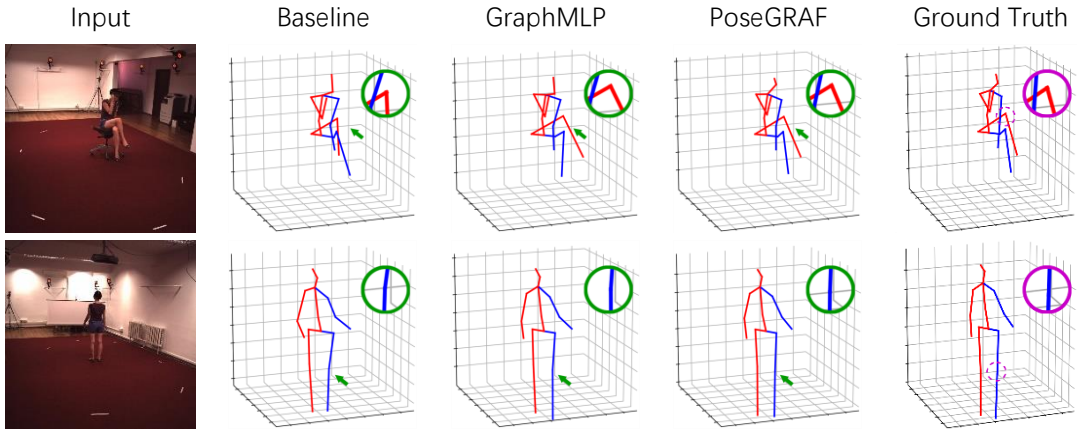

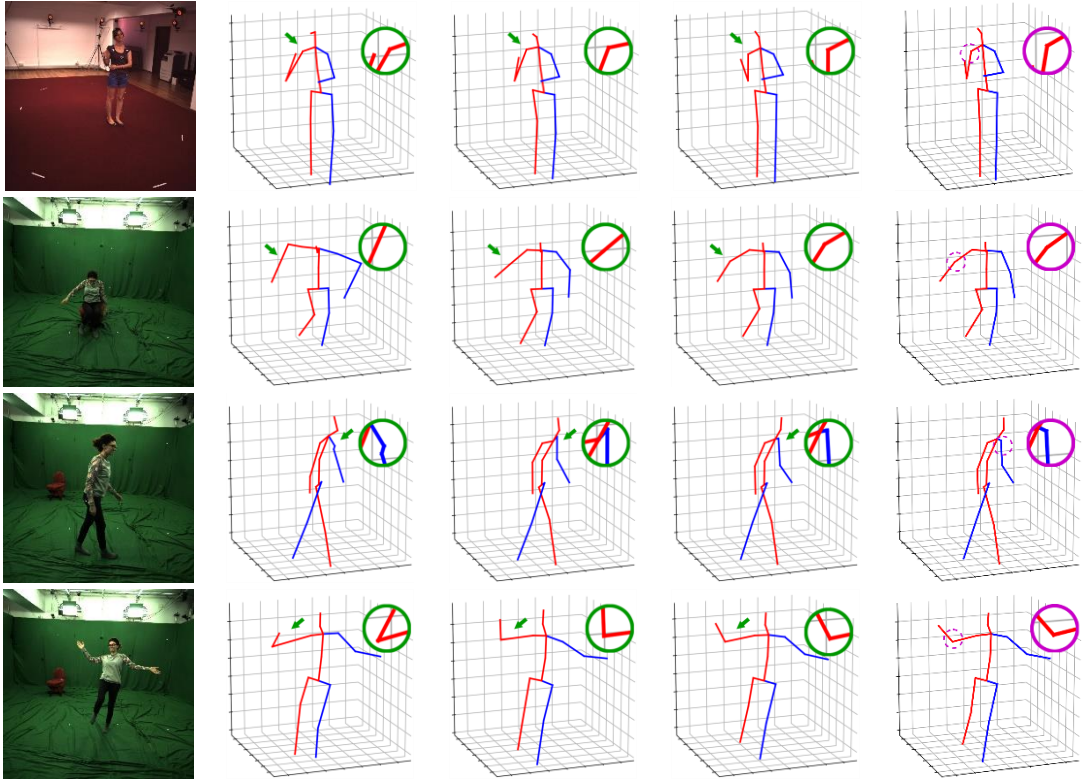

Fig. 4. 3D pose estimation visualizations for Human3.6M (top three rows) and MPI-INF-3DHP (bottom three rows) datasets.

### 4.5 Qualitative results on video in-the-wild

To evaluate the model's robustness in real-world open environments, this study designs a cross-domain generalization validation protocol to address dual challenges of unknown camera parameters and complex dynamic scenarios. The system implementation framework comprises three stages: a human detection module localizes target subjects in video frames, a high-precision 2D pose estimator based on HRNet extracts spatiotemporal keypoint sequences, and the pre-trained PoseGRAF is transferred to diverse motion videos for end-to-end 3D reconstruction. The test set covers highly challenging pose sequences including high-intensity fitness, gymnastic movements, and basketball playing. Quantitative visualization results (see Fig. 5) demonstrate that, under completely unseen scenarios, our method consistently outputs anatomically plausible 3D human poses in 3D space.

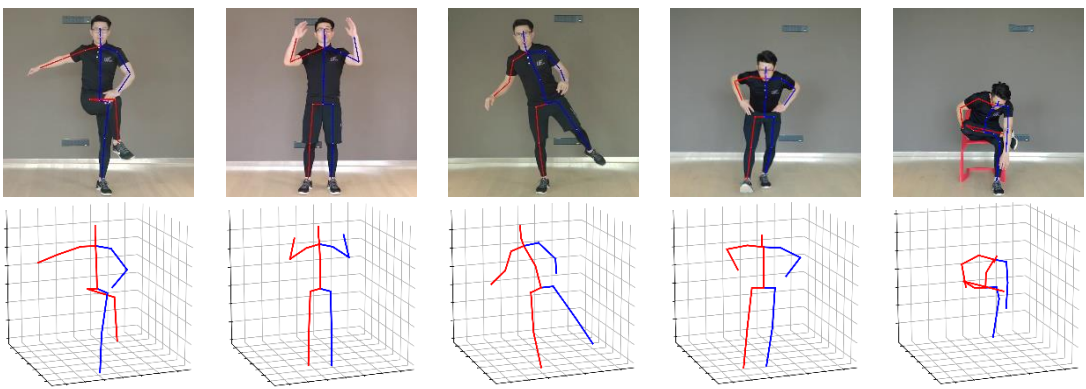

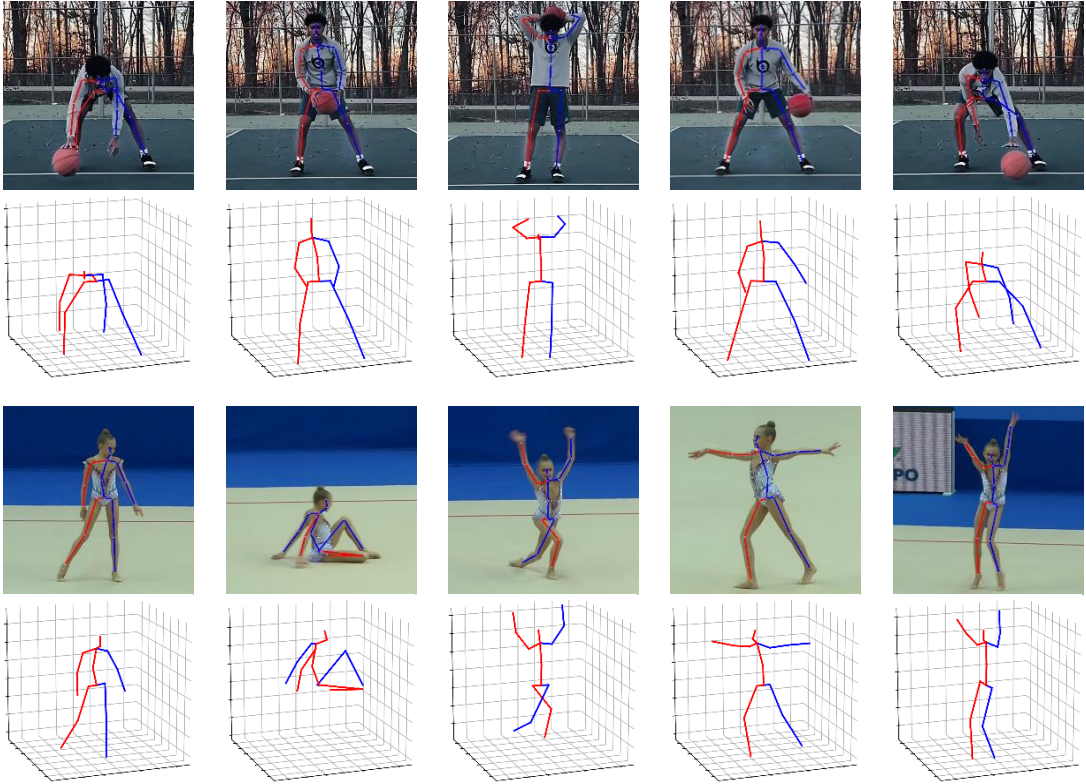

Fig. 5. Qualitative results of our method for in-the-wild videos.

## 5  Conclusion

This paper proposes PoseGRAF, a geometry-enhanced graph learning framework, addressing the limitations of monocular 3D human pose estimation caused by over-reliance on 2D joint coordinates and insufficient utilization of bone directions and joint angles. By constructing a geometry-enhanced graph structure that unifies the encoding of bone directions and joint angles, integrating graph convolution modules to capture skeletal spatial correlations, and introducing an attention-driven dynamic feature fusion mechanism to adaptively consolidate global joint positions with local geometric features, our method effectively mitigates prediction bias in occluded scenarios. On the Human3.6M and MPI-INF-3DHP datasets, PoseGRAF significantly outperforms state-of-the-art methods and demonstrates robust generalization capabilities in in-the-wild video testing. Future work will explore temporal kinematic constraints and lightweight deployment to advance real-time applications such as medical rehabilitation.